\pdfoutput=1

\documentclass[11pt]{article}

\usepackage[]{ACL2023}

\usepackage{times}
\usepackage{latexsym}

\usepackage[T1]{fontenc}

\usepackage[utf8]{inputenc}

\usepackage{microtype}

\usepackage{inconsolata}

\usepackage{graphicx}
\graphicspath{ {./images/} }

\usepackage{booktabs}
\usepackage{amsmath}
\usepackage{enumitem}
\usepackage{xcolor}
\usepackage{xspace}
\usepackage{makecell}

\usepackage[disable]{todonotes}
\makeatletter
\newcommand*\iftodonotes{\if@todonotes@disabled\expandafter\@secondoftwo\else\expandafter\@firstoftwo\fi}
\makeatother

\newcommand{\note}[4][]{{\todo[author=#2,color=#3,size=\scriptsize,fancyline,caption={},#1]{#4}}}

\newcommand{\jason}[2][]{\note[#1]{jason}{green!40}{#2}}

\newcommand{\cutforspace}[2][]{\note[#1]{cut for space}{gray}{#2}}

\usepackage{soul}
\usepackage{csquotes}

\definecolor{lightpurple}{rgb}{0.9, 0.8, 1}
\definecolor{lightred}{rgb}{1, 0.8, 0.8}
\definecolor{lightblue}{rgb}{0.8,0.9,1}

\DeclareRobustCommand{\hlpurple}[1]{{\sethlcolor{lightpurple}\hl{#1}}}
\DeclareRobustCommand{\hlgreen}[1]{{\sethlcolor{green}\hl{#1}}}
\DeclareRobustCommand{\hlred}[1]{{\sethlcolor{lightred}\hl{#1}}}

\DeclareRobustCommand{\hlblue}[1]{{\sethlcolor{lightblue}\hl{#1}}}

\title{Fine-Tuning LLMs with Fine-Grained Human Feedback on Text Spans}

\author{
Sky CH-Wang$^{\bullet}$\quad
Justin Svegliato$^{\circ}$\quad
Helen Appel$^{\circ}$\quad
Jason Eisner$^{\ominus}$\\
$^{\bullet}$Columbia University\quad
$^{\circ}$Microsoft\quad
$^{\ominus}$Johns Hopkins University\quad\\
\texttt{skywang@cs.columbia.edu}, \texttt{\{jsvegliato, helenappel\}@microsoft.com}, \texttt{jason@cs.jhu.edu}
}

\begin{document}
\maketitle
\begin{abstract}
We present a method and dataset for fine-tuning language models with preference supervision using feedback-driven improvement chains.
Given a model response, an annotator
provides fine-grained feedback by marking ``liked'' and ``disliked'' spans 
and specifying what they liked or disliked about them.
The base model then rewrites the disliked spans accordingly, proceeding from left to right,
forming a sequence of incremental improvements. 
We construct preference pairs for 
direct alignment
from each adjacent step in the chain, enabling the model to learn from localized, targeted edits.
We find that our approach outperforms direct alignment methods based on standard A/B preference ranking or full contrastive rewrites, demonstrating that structured, revision-based supervision leads to more efficient and effective preference tuning.
\end{abstract}

\section{Introduction}\label{sec:intro}

\begin{figure*}[ht]
  \small
  \centering
  \includegraphics[width=0.9\linewidth,height=3cm]{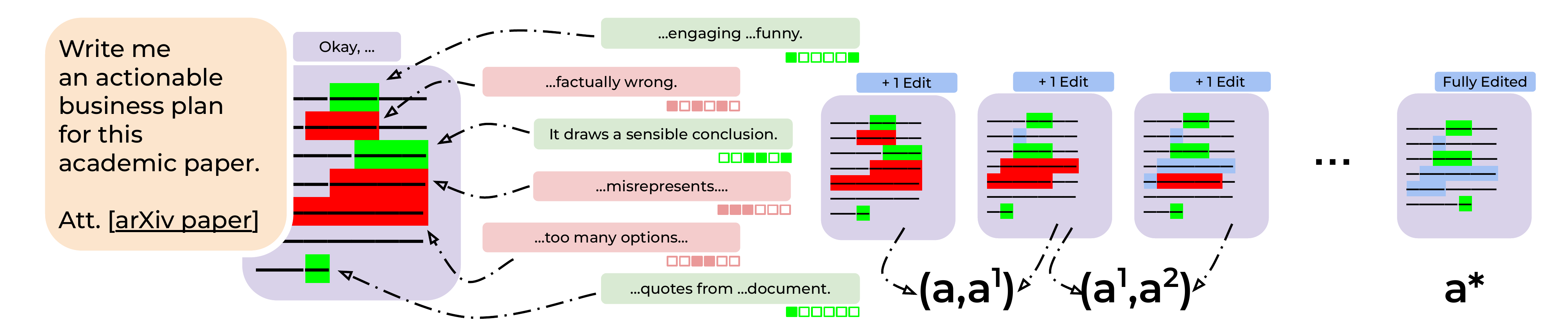}
  \caption{An overview of our critique-guided improvement chain framework. A \hlpurple{model response} is annotated with span-level feedback (left) in the form of \hlgreen{like} and \hlred{dislike} highlights---each accompanied by labeled rationale checkboxes---and then revised \textit{step-by-step} (right) into a final fully-edited response ($a^*$). Each \textit{adjacent pair} ($a^{i},a^{i+1}$) in the improvement sequence is used as a preference pair for direct alignment. \hlblue{Blue} indicates a rewrite.
  }
\label{fig:pipeline}
\end{figure*}

Large language models (LLMs) achieve strong performance across natural language tasks by fine-tuning on human preferences \cite{ouyang2022training, bai2022constitutional}, also known as reinforcement learning from human feedback (RLHF). Direct alignment methods such as direct preference optimization (DPO) \cite{rafailov2023direct}
tune the LLM directly on preference pairs.\footnote{Alternative methods tune it using a reward model trained to predict preferences on such pairs \cite{christiano2017deep}.}
To construct a preference pair, it is typical to sample two random responses from a model and then ask humans to rate which one is better \cite{ouyang2022training}. Unfortunately, these judgments can be difficult to make and noisy in nature \cite{chowdhury2024provably}, as it is rare for either response to dominate the other in the sense of being better in all aspects. 
We propose an alternative human feedback framework in which sampled responses are \textit{revised} to produce preference pairs with \textit{clearer} and \textit{more meaningful} preference distinctions. In our framework, humans are used not to \textit{compare} responses but rather to \textit{indicate opportunities for revision} (Figure \ref{fig:pipeline}). 

We introduce a novel \textit{dataset} in which human annotators highlight specific spans that they ``like'' or ``dislike'' in (long) model responses.
They also indicate \textit{why} they liked or disliked each span, using a taxonomy we propose. This provides far more information than a one-bit A/B preference ranking, while adding negligible annotation overhead (\S\ref{sec:annotation-time}).

Our novel dataset could help improve an LLM in many ways, such as by providing multi-objective and localized reward signals \cite{wu2023fine}. 
However, in this paper, we go beyond extracting reward signals from the span-level annotations. Instead, we prompt the \textit{original response model} to examine all the span-level feedback jointly and make targeted \emph{revisions} to address it, yielding improved responses that dominate the originals and can be used as preference pairs for direct alignment methods.
We test how different strategies for building preference pairs with this feedback---\textit{single-span} rewrites, full \textit{all-at-once} rewrites, and \textit{cumulative step-by-step} rewrites---affect preference learning and downstream models under direct alignment.  We find that cumulative rewrites yield the \textit{strongest} models and may improve sample efficiency.

If the LLM represents an infinite population of responses, our approach to improving it resembles the \emph{Lamarckian} theory of evolution.  Through revision, each response \emph{adapts} to its environment of human judges; fine-tuning then ensures that these beneficial adaptations are inherited by the next generation.  Standard RLHF uses the less efficient \emph{Darwinian} mechanism of selecting for adaptive traits that \emph{already} appear in the population through natural variation.\footnote{In this analogy, traits are not passed into the next generation of LLM responses by producing descendants of individual ``organisms'' as in evolution.
Instead, the LLM parameters are tuned to place more probability on either the adapted responses (Lamarckian) or the adaptive responses (Darwinian).  This tends to increase the prevalence of their traits in the next generation by also raising the probability of similar responses.}
We find that our Lamarckian approach leads to better alignment with less training.

\section{Related Work}

\paragraph{Feedback Formats.} 
Traditional preference-tuning approaches often rely on simple A/B comparisons of complete responses \cite{ouyang2022training}: human or LLM judges \cite{pmlr-v235-cui24f} select the preferred response without providing any rationale. More recent efforts have sought richer supervision data.  Human or LLM judges may deliver 
fine-grained and/or multidimensional feedback
on individual responses
\cite{wu2023fine, wang-etal-2024-helpsteer,ye-etal-2025-improving}, preference pairs \cite{pmlr-v235-cui24f,just2025datacentric}, or revision pairs \cite{guo2023leveraging}.
Feedback on real chatbot responses can also be imputed from the conversational reactions that they elicit \cite{lin2024interpretable, shi2024wildfeedback}.

More recently, another paradigm invites domain experts to \emph{directly edit} generated responses \cite{chakrabarty2025ai}.
In our evolution analogy, this corresponds to \emph{intelligent design}.  But when humans are the editors, it often
imposes a high cognitive load and struggles to scale.
By contrast, in our lower-effort Lamarckian method, human reviewers simply highlight favored and disfavored segments of text and select and/or write 
brief explanations.  The responses then ``adapt'' to this environmental feedback.

\paragraph{Self-training.}  Since \citet{huang2022selfimprove}, many papers 
obtain high-quality responses through some expensive multi-step LLM workflow, and then use those responses for supervised fine-tuning (which teaches the LLM to generate them in a \emph{single} step) or for preference optimization \citep[which teaches the LLM to prefer them to its current single-step responses:][]{dong2024selfboosting,d2024anchored}.
Our approach fits into this line of work, but we aim for \emph{human alignment} by identifying a step of our multi-step workflow---feedback on spans---where human supervision is relatively cheap.\footnote{In future work, we would like to evaluate the impact on cost and alignment quality of automating this step.}  
The final responses are still generated by the current LLM and are therefore, we hope, achievable by fine-tuning.
This is reminscent of supervising machine translation through ``hope'' responses that are generated by the current machine translation system, but under a modified objective that encourages them to be similar to a human-provided reference response \cite{JMLR:v13:chiang12a}.

\paragraph{Preference Optimization.}
Direct Preference Optimization (DPO) \cite{rafailov2023direct} 
enables stable fine-tuning from preference pairs without learning an explicit reward function.  However, its effectiveness depends heavily on how preference data are constructed.
Recent work explores improvements including curriculum learning \cite{pattnaik-etal-2024-enhancing}, optimal preference pair selection \cite{xiao2025finding}, and alternative loss formulations \cite{pal2024smaug}.

Since our preference pairs differ only by small local edits, we experiment with various training losses for direct alignment \cite{d2024anchored, pal2024smaug}
that are designed to handle highly similar preference pairs.  Several studies have reported that DPO can fail catastrophically in this setting, for example due to ``likelihood displacement'' \cite{razin2025unintentional}.

\section{Data}\label{sec:data}\label{sec:annotation-time}

\begin{table}[t]
\small
\centering
\begin{tabular}{lccccc}
\toprule
\textbf{Domain} & $N$ & $H^+$ & $H^-$ & $\Sigma A$ & $\#w$ \\
\midrule
Yelp       & 86 & 45 & 397 & 1589 & 36k \\
News       & 81 & 49 & 379 & 1693 & 31k \\
Wikipedia  & 67 & 24 & 334 & 1375 & 28k \\
arXiv      & 43 & 27 & 193 & 863 & 19k \\
\midrule
\textbf{Total} & 277 & 145 & 1303 & 5520 & 115k \\
\bottomrule
\end{tabular}
\caption{Annotation statistics by domain. $N$ is the number of responses annotated. $H^+$ and $H^-$ are the number of span-level like and dislike highlights. $\Sigma A$ is the total number of categorical attributes selected (checkboxes ticked), and $\#w$ is the total number of words highlighted, as calculated via whitespace tokenization.}
\label{tab:annotation_stats}
\end{table}

\paragraph{Response Generation.}
Our study focuses on \textit{long-form in-context generation}---e.g., the generation stage of retrieval-augmented generation.  The length and informational density of typical responses in this setting make it natural for users to form varying judgments about different parts of the output. This creates a natural fit for localized feedback, yet traditional preference learning methods offer only global comparisons, making it difficult to isolate what exactly should be improved.

To construct a set of model responses for human annotation, we first sample documents from four diverse long-context domains: (1) Yelp reviews \cite{zhang2015character},
(2) News articles from Multi-News \cite{fabbri-etal-2019-multi},
(3) Wikipedia pages \cite{wikidump2024}, and
(4) AI-related arXiv papers.\footnote{\url{https://hf.co/datasets/jamescalam/ai-arxiv2}}
For each document, we use \texttt{GPT-4.1} \cite{openai2025gpt41} to generate $5$ challenging user queries---designed to elicit responses requiring reasoning, synthesis, or subjective judgment---that could plausibly retrieve the document in a retrieval-augmented generation setup. For each query, we sample 2 independent responses from \texttt{Llama-3.1-8B-Instruct} \cite{grattafiori2024llama} with $0.8$ temperature and $0.95$ top-$p$ sampling, using a fixed prompt template that includes the source document. Our prompt templates are provided in Appendix~\ref{appendix:prompts} with example queries and responses.\cutforspace{We then collect span-level human feedback and preference annotations on these responses.}

\paragraph{Annotation.} 
We began with a pilot study in which 3 annotators were presented with the query, source document, and the 2 model responses, and asked to consider what they would have wanted to see in a good response to the query, given the document. Annotators \textit{highlighted} spans they liked or disliked in 100 model responses to 50 queries
and gave a brief \textit{explanation} for each marked span as a natural-language phrase or sentence.
We conducted thematic analysis following \citet{braun2006using}’s 6-phase approach to derive a taxonomy of common preference attributes (20 like/19 dislike).
With the full taxonomy defined (Appendix~\ref{appendix:feedback}), we augmented the annotation interface to let annotators provide categorical feedback by selecting attribute checkboxes for each highlighted span.\looseness=-1

We then presented 4 new annotators with pairs of Llama responses.  They (1) highlighted \textit{spans} they liked or disliked in each response, (2) marked span attributes
(plus optional free-text feedback), 
and (3) made an \textit{A/B preference judgment} with a brief explanation. Dataset statistics are in Table~\ref{tab:annotation_stats}.  On average, they marked 0.5 like and 4.7 dislike spans per response, with 3.8 attributes per span.  Our instructions to annotators are in Appendix \ref{appendix:annotation}.

To measure inter-annotator agreement, 100 items were annotated by all annotators.\jason{I don't think this was just to measure agreement---isn't this all of the items?  Can we say that the same 50 pairs were annotated by the 4 new annotators?}
Agreement on A/B preference rankings was moderate (Fleiss $\kappa=0.47$), reflecting the difficulty of such judgments (see \S\ref{sec:intro}). Agreement on exact spans and their attributes was negligible.
This is unsurprising for a fine-grained subjective feedback task
and does not mean that annotators actively disagreed: an annotator was not asked to label \emph{every} good and bad aspect of the response and might omit many labels from other annotators\footnote{Or deviate from them, conveying very similar feedback with slightly different span boundaries or checkboxes.} that they would nonetheless endorse as reasonable if asked.\jason{we should ask!! can do it soon and report in camera-ready} 
Thus, we treat the observed variability as additional information rather than noise: it reveals the breadth of human preferences that the LLM should accommodate. Because each decision is paired with checkbox rationales, the dataset still provides richly grounded, interpretable supervision from which models can learn.\jason{I cut this line: "Examples used in training are singly annotated", which might suggest that we keep only 1 annotator for each response and throw out the other 3 (which would reduce us to at most 100 responses, not 277, so I think we don't do that, right?).}\looseness=-1

\paragraph{Improvement Sequence Generation.} 
To construct a \textit{step-wise improvement sequence} from a response annotated by a particular annoator, we prompt the original response model (\texttt{Llama-3.1-8B-\allowbreak Instruct}) with four inputs: the initial response, this annotator's complete feedback on liked and disliked spans, the source document, and an instruction explicitly requesting a sequence of incremental edits---where each step addresses \textit{exactly one} disliked span, proceeding from left to right.  The goal is to produce a chain of responses in which each adjacent pair differs by a single, targeted revision.  An overview of the framework is in Figure~\ref{fig:pipeline}, with full prompts in Appendix~\ref{appendix:prompts}.

Note that the annotator is free to mark a narrow span (e.g., highlighting an imprecise word), and the LLM is equally free to edit beyond the span's boundaries if needed to fix the problem (e.g., rethinking the whole clause, or changing the word but also adjusting the adjacent context to maintain coherence and fluency).  We do apply a Levenshtein distance heuristic\jason{details needed} to ensure structural compliance: sequences that fail this check (e.g., due to multi-edit steps or non-contiguous rewrites) are discarded and regenerated.  This process yields 277 valid improvement sequences for a total of 1303 unique improvement steps.\jason{I'm confused.  The response to a reviewer said that we got down from 400 to 277 through this quality control procedure, but if the discarded ones were regenerated, how did that reduce the number?} 

\paragraph{Annotation Time.} 
The time and financial cost of the LLM calls was negligible compared to the human annotation.
We logged how long annotators spent on each  item, omitting outliers beyond 1.5 times the inter-quartile range from the median to reduce the influence of breaks and distractions.
Notably, annotating a \textit{pair} of responses with our full protocol (455 seconds) 
was only 9\% slower than simple A/B preference ranking (419 seconds). 
This low overhead likely results from A/B ranking \textit{already} being difficult, especially in our task. Annotators had to carefully read the deliberately challenging query and both long responses and implicitly reason about their preferences. Prior work similarly finds that making such implicit rationales explicit adds little additional burden \cite{zaidan-etal-2007-using}.\looseness=-1 

Notice that A/B ranking produces only 1 pair for preference tuning, while our protocol can produce $1+4.7+4.7 = 10.4$ pairs on average (the A/B comparison, plus one synthetic pair per dislike span on each of A and B).  This makes our protocol more than $9\times$ faster \emph{per pair}.\jason{but we discarded some pairs?  need to correct for that}\footnote{We do caution against simply counting pairs.  Perhaps not all types of pairs are equally likely for training (and our method did not even train on the A/B comparison).  But overall, Table~\ref{tab:main_results} shows that we improved ELO from 1612 to 1634 with only a 9\% increase in human annotation time.}  

In future work, it would be worth investigating reduced-cost protocols.  Was it necessary to require the comparison of two responses (to force careful reading), or could the annotator simply read and annotate a single response?  What if they only marked the 1 or 2 spans that they felt most strongly about?  What if they skipped the like spans (which were not rewritten), or the span attributes (so that the LLM revision model had to guess what was wrong with the span)?  Finally, could spans be marked by the actual user who solicited the response, as a richer alternative to thumbs-up/thumbs-down feedback?  

\section{Experiments}

With our dataset, we preference-tune \texttt{Llama-3.1\allowbreak-8B-Instruct}, seeking to understand how different forms of feedback-derived preference pairs affect model learning and performance. Specifically---where \((x, y)\) denotes a pair where \(y\) is the \textit{winner} response and \(x\) is the \textit{loser}---we compare:

\begin{itemize}[noitemsep,topsep=3pt,leftmargin=.1in]
    \item \textbf{Preference Pairs} \((a, b)\) where \(a\) and \(b\) are two ranked Llama responses to the same prompt.
    \item \textbf{First Edits}
    $(a^0,a^1)$ where $a^1$ is a single edit that improves just the \textit{first} disliked span. 
    \item \textbf{Full Rewrites} 
    $(a^0,a^*)$
    where \(a^*\) is the final \textit{cumulative rewrite} of the original response \(a^0\).
    \item \textbf{Stepwise Edits} \((a^{i}, a^{i+1})\) where \(a^{i+1}\) is a stepwise revision of \(a^{i}\), generated to address a \textit{single} span-level critique. Each pair corresponds to two adjacent steps in the improvement sequence.
\end{itemize}

\noindent We also evaluate performance against two baselines: the base model (\texttt{Llama-3.1-8B-\allowbreak Instruct}) and an SFT baseline ($a^*$ SFT), in which the base model is fine-tuned on fully-improved $a$ responses. 

Following standard post-training protocols, all preference-tuning methods are applied on top of the $a^*$ SFT model, with the exception of the $(a, b)$ setting (in which $a^*$ is unavailable).
We treat the training loss function as a hyperparameter, considering DPO \cite{rafailov2023direct}, DPO-Positive \cite{pal2024smaug}, APO-zero, and APO-down \cite{d2024anchored}.   
That is, for each \textit{preference pair construction method}, we fine-tune with each training loss and evaluate the method on test data using the fine-tuned model that performed best on separate validation data.  
This procedure selected APO-down---well-suited for contrastive preference data---for all methods except $(a,b)$, for which DPO was better.
Hyperparameters for training configurations are in Appendix~\ref{appendix:hyperparams}.

A possible advantage of Stepwise Edits is that it produces more training signal by converting each annotated response into \textit{multiple} preference pairs $(a^i, a^{i+1})$.
To assess how much of this method's benefit comes from the increased  \emph{number} of  pairs vs.\@ from their \emph{diversity} (relative to $(a^0,a^1)$) or \emph{minimality} (relative to $(a^0,a^*)$), we also try down-sampling these preference pairs $(a^i, a^{i+1})_{\text{ds}}$ to match the number to other methods.

We evaluate each model based on its Elo score (Appendix~\ref{appendix:hyperparams}), computed from 263 pairwise comparisons
judged by 4 human (\textbf{ELO\textsubscript{H}}) annotators on responses generated with a temperature of $0.8$ and top-$p$ sampling of $0.95$ on a held-out set of 40 prompts. We also report automated annotator (\textbf{ELO\textsubscript{M}}) results
on responses to 138 held-out prompts (2898 pairwise comparisons), using the \texttt{alpaca\_eval\_gpt4} evaluator in \citet{alpaca_eval} with the highest correlation to human judgments.

\begin{table}[t]
    \centering
    \small
    \begin{tabular}{lrlrl}
    \toprule
    \textbf{Method} & \multicolumn{2}{c}{\textbf{ELO\textsubscript{H}}} & \multicolumn{2}{c}{\textbf{ELO\textsubscript{M}}} \\
    \midrule
    \texttt{base}                       & 1383 & \tiny{(-305,-71)} & 1355 & \tiny{(-314,-223)} \\
    $a^*$ SFT                           & 1377 & \tiny{(-310,-70)} & 1353 & \tiny{(-313,-234)} \\
    \((a, b)\)                          & 1465 & \tiny{(-247,-4)} & 1376 & \tiny{(-291,-207)} \\
    \((a^0, a^1)\)                         & 1525 & \tiny{(-183,+52)} & 1435 & \tiny{(-232,-147)} \\
    \((a^0, a^*)\)                        & 1612 & \tiny{(-155,+52)} & 1617 & \tiny{(-57,+31)} \\
    \midrule
    $(a^i, a^{i+1})_{\text{ds}}$        & 1620 & \tiny{(-120,+105)} & 1602 & \tiny{(-76,+28)} \\
    \((a^i, a^{i+1})\)                  & \textbf{1634} & & \textbf{1629} & \\
    \bottomrule
    \end{tabular}
    \caption{Comparative quality of the 7 models' responses using ELO scores under human judgments (\textbf{ELO\textsubscript{H}}) and \texttt{alpaca\_eval\_gpt4} judgments (\textbf{ELO\textsubscript{M}}). 
    We show a bootstrap 95\% confidence interval for each model's difference from the best-performing model.
    \((a^i, a^{i+1})\) benefits from training on all 1303 stepwise preference pairs; all other methods generate only 277 pairs. 
    }
    \label{tab:main_results}
\end{table}

Table~\ref{tab:main_results} shows that our approach
significantly outperforms standard A/B preference ranking. Given the negligible annotation overhead relative to A/B preferences, this highlights the efficiency of our framework in gathering actionable and effective preference data from human annotators. 
When analyzing how best to utilize this feedback for direct alignment, even preference pairs derived from \textit{single-step edits}---targeted revisions addressing a single highlighted span---outperform A/B preference data.
Further gains are achieved through \textit{cumulative rewriting}, where iterative revisions produce a final response that is often better than any model output sampled directly. Using the original and final responses in this sequence as a preference pair yields a large improvement in model quality.\footnote{Such a preference pair could be generated more cheaply in a single step, rather than through a series of intermediate steps.}
Finally, incorporating the \textit{intermediate} steps as additional training pairs---teaching the model to prefer each version over its predecessor---provides a small but not statistically significant boost.
Overall, these results suggest that our annotation framework---centered on span-level feedback and iterative revision---provides a more effective/scalable way to elicit high-quality preference data, enabling better model alignment with minimal additional annotation effort.

\vspace{-1pt}
\section{Conclusion}
\vspace{-1pt}

We offer a low-overhead framework for preference-tuning that uses span-level feedback and stepwise rewrites to generate structured improvement sequences. By constructing preference pairs from adjacent edits in these sequences, our method enables models to learn from localized, fine-grained supervision. Experiments show that this fine-grained approach to preference optimization outperforms existing alignment techniques, highlighting the value of \textit{targeted feedback} and \textit{minimally contrastive} training signals in shaping model behavior. 

We will release our annotation tool and dataset.  Our novel feedback taxonomy is provided in Appendix~\ref{appendix:feedback}.

\section*{Limitations}

\paragraph{Alignment Objective.}
How to train on revision data in a principled way is an open question.  For convenience, we experimented with existing preference optimization methods, but it might be more appropriate to design new ones.  The standard methods are based on the notion that when humans stochastically rank $a \prec b$, this provides additional evidence that $\mathrm{reward}(a) < \mathrm{reward}(b)$.  In contrast, our revision process may tend to improve the reward of a response, but we do not have a way of estimating by how much.\footnote{Without asking human or LLM judges.}  It is not even clear that reward does improve at each revision step---perhaps the intermediate steps in $(a^0, a^1, \ldots a^*)$ have lower reward (and should have lower probability) because they are internally inconsistent in style or goals.\jason{although if the reference model \emph{knows} they're inconsistent, they can still have lower prob with higher reward.  I was thinking here about the DPO suffix issue that I posted on Slack. 
 Requires more thought.}\footnote{We also note a technical roadblock: we cannot compute the closed-form probability of generating a certain revised response during training.  That rules out clipped importance sampling methods like GRPO, which require this proposal probability for use as a denominator.}
Empirically, we observe that some standard methods fail: when using DPO on stepwise-edit-constructed preference pairs, training can exhibit catastrophic degeneration.
It remains an open question how to design direct alignment methods that faithfully reflect the structure and semantics of revision-constructed preference data, while preserving stability and effectiveness during training.

\paragraph{Frontier Models.} Our experiments are with an 8B-parameter model.  It is unknown whether this workflow would also be able to improve today's frontier models.

\paragraph{Removing the Humans.}
In principle, the span-level feedback could itself be provided by the LLM.  We have not yet experimented with this workflow.  At that point, we would be using the LLM to reflect on its own answers \emph{and} improve them, and then training it to produce or prefer the improved answers.  Recent work on standard RLHF does increasingly substitute high-quality LLM preferences for human preferences \cite[e.g.,][]{pmlr-v235-cui24f}.
Compared to humans, AI-generated feedback is more scalable, easier to collect, and significantly cheaper.  On the other hand, retaining some response-level feedback from actual humans might be important to achieve or at least evaluate alignment with actual human values.

\paragraph{Annotation Instructions.}
Our annotation guidelines did not tell annotators how many spans to mark per response (or how many explanations to mark per span).  Best practices for such guidelines or incentives remain to be worked out.  (However, the $(a^i, a^{i+1})_{\text{ds}}$ result in Table~\ref{tab:main_results} suggests that marking even a single span per response can work well.)

\paragraph{Other Settings.}
Our taxonomy that guided the human feedback was developed specifically for our RAG setting.  Feedback on tasks such as social dialogue, advice, therapy, tutoring, or code generation---to give a few examples---would presumably look quite different; those settings would require developing new taxonomies.  Our annotation interface and revision prompt might also not extend to a setting like code, where perhaps feedback should no longer be attached to textual spans.

\bibliography{acl2023}

@string{acl = {Association for Computational Linguistics}}

@string{anth = {https://aclanthology.org/}}

@string{naacl:2024:long = {Proceedings of the 2024 Conference of the North American Chapter of the Association for Computational Linguistics: Human Language Technologies (Volume 1: Long Papers)}}

@inproceedings{wang-etal-2024-helpsteer,
  title = {{H}elp{S}teer: Multi-attribute Helpfulness Dataset for {S}teer{LM}},
  author = {Wang, Zhilin  and
      Dong, Yi  and
      Zeng, Jiaqi  and
      Adams, Virginia  and
      Sreedhar, Makesh Narsimhan  and
      Egert, Daniel  and
      Delalleau, Olivier  and
      Scowcroft, Jane  and
      Kant, Neel  and
      Swope, Aidan  and
      Kuchaiev, Oleksii},
  editor = {Duh, Kevin  and
      Gomez, Helena  and
      Bethard, Steven},
  booktitle = naacl:2024:long,
  month = jun,
  year = {2024},
  address = {Mexico City, Mexico},
  publisher = acl,
  url = anth # {2024.naacl-long.185},
  doi = {10.18653/v1/2024.naacl-long.185},
  pages = {3371--3384}
}

@string{acl:2019:1 = {Proceedings of the 57th Annual Meeting of the Association for Computational Linguistics}}

@inproceedings{fabbri-etal-2019-multi,
  title = {Multi-News: A Large-Scale Multi-Document Summarization Dataset and Abstractive Hierarchical Model},
  author = {Fabbri, Alexander  and
      Li, Irene  and
      She, Tianwei  and
      Li, Suyi  and
      Radev, Dragomir},
  editor = {Korhonen, Anna  and
      Traum, David  and
      M{\`a}rquez, Llu{\'\i}s},
  booktitle = acl:2019:1,
  month = jul,
  year = {2019},
  address = {Florence, Italy},
  publisher = acl,
  url = anth # {P19-1102},
  doi = {10.18653/v1/P19-1102},
  pages = {1074--1084}
}

@string{naacl:2007:1 = {Human Language Technologies 2007: The Conference of the North {A}merican Chapter of the Association for Computational Linguistics; Proceedings of the Main Conference}}

@inproceedings{zaidan-etal-2007-using,
  title = {Using {``}Annotator Rationales{''} to Improve Machine Learning for Text Categorization},
  author = {Zaidan, Omar  and
      Eisner, Jason  and
      Piatko, Christine},
  editor = {Sidner, Candace  and
      Schultz, Tanja  and
      Stone, Matthew  and
      Zhai, ChengXiang},
  booktitle = naacl:2007:1,
  month = apr,
  year = {2007},
  address = {Rochester, New York},
  publisher = acl,
  url = anth # {N07-1033},
  pages = {260--267}
}

@article{ouyang2022training,
  title = {Training Language Models to Follow Instructions with Human Feedback},
  author = {Ouyang, Long and Wu, Jeffrey and Jiang, Xu and Almeida, Diogo and Wainwright, Carroll and Mishkin, Pamela and Zhang, Chong and Agarwal, Sandhini and Slama, Katarina and Ray, Alex and others},
  journal = {Advances in Neural Information Processing Systems},
  volume = {35},
  pages = {27730--27744},
  year = {2022},
  url = {https://arxiv.org/abs/2203.02155}
}

@article{wu2023fine,
  title = {Fine-Grained Human Feedback Gives Better Rewards for Language Model Training},
  author = {Wu, Zeqiu and Hu, Yushi and Shi, Weijia and Dziri, Nouha and Suhr, Alane and Ammanabrolu, Prithviraj and Smith, Noah A and Ostendorf, Mari and Hajishirzi, Hannaneh},
  journal = {Advances in Neural Information Processing Systems},
  volume = {36},
  pages = {59008--59033},
  year = {2023},
  url = {https://arxiv.org/abs/2306.01693}
}

@article{bai2022constitutional,
  title = {Constitutional {AI}: Harmlessness from {AI} Feedback},
  author = {Bai, Yuntao and Kadavath, Saurav and Kundu, Sandipan and Askell, Amanda and Kernion, Jackson and Jones, Andy and Chen, Anna and Goldie, Anna and Mirhoseini, Azalia and McKinnon, Cameron and others},
  journal = {Computing Research Repository (CoRR)},
  volume = {arXiv:2212.08073},
  year = {2022},
  url = {https://arxiv.org/abs/2212.08073}
}

@article{rafailov2023direct,
  title = {Direct Preference Optimization: Your Language Model is Secretly a Reward Model},
  author = {Rafailov, Rafael and Sharma, Archit and Mitchell, Eric and Manning, Christopher D and Ermon, Stefano and Finn, Chelsea},
  journal = {Advances in Neural Information Processing Systems},
  volume = {36},
  pages = {53728--53741},
  year = {2023},
  url = {https://arxiv.org/abs/2305.18290}
}

@article{christiano2017deep,
  title = {Deep Reinforcement Learning from Human Preferences},
  author = {Christiano, Paul F and Leike, Jan and Brown, Tom and Martic, Miljan and Legg, Shane and Amodei, Dario},
  journal = {Advances in Neural Information Processing Systems},
  volume = {30},
  year = {2017},
  url = {https://arxiv.org/abs/1706.03741}
}

@article{braun2006using,
  title = {Using Thematic Analysis in Psychology},
  author = {Braun, Virginia and Clarke, Victoria},
  journal = {Qualitative Research in Psychology},
  volume = {3},
  number = {2},
  pages = {77--101},
  year = {2006},
  publisher = {Taylor \& Francis},
  url = {https://doi.org/10.1191/1478088706qp063oa}
}

@article{grattafiori2024llama,
  title = {The {Llama 3} Herd of Models},
  author = {Grattafiori, Aaron and Dubey, Abhimanyu and Jauhri, Abhinav and Pandey, Abhinav and Kadian, Abhishek and Al-Dahle, Ahmad and Letman, Aiesha and Mathur, Akhil and Schelten, Alan and Vaughan, Alex and others},
  journal = {Computing Research Repository (CoRR)},
  volume = {arXiv:2407.21783},
  year = {2024},
  url = {https://arxiv.org/abs/2407.21783}
}

@misc{openai2025gpt41,
  author = {OpenAI},
  title = {Introducing {GPT}-4.1 in the {API}},
  year = {2025},
  month = {apr},
  url = {https://openai.com/index/gpt-4-1/},
  note = {Accessed: 2025-04-24}
}

@inproceedings{pmlr-v235-cui24f,
  title = {{ULTRAFEEDBACK}: Boosting Language Models with Scaled {AI} Feedback},
  author = {Cui, Ganqu and Yuan, Lifan and Ding, Ning and Yao, Guanming and He, Bingxiang and Zhu, Wei and Ni, Yuan and Xie, Guotong and Xie, Ruobing and Lin, Yankai and Liu, Zhiyuan and Sun, Maosong},
  booktitle = {Proceedings of the 41st International Conference on Machine Learning},
  pages = {9722--9744},
  year = {2024},
  editor = {Salakhutdinov, Ruslan and Kolter, Zico and Heller, Katherine and Weller, Adrian and Oliver, Nuria and Scarlett, Jonathan and Berkenkamp, Felix},
  volume = {235},
  series = {Proceedings of Machine Learning Research},
  month = {21--27 Jul},
  publisher = {PMLR},
  url = {https://proceedings.mlr.press/v235/cui24f.html}
}

@article{d2024anchored,
  author = {D'Oosterlinck, Karel and Xu, Winnie and Develder, Chris and Demeester, Thomas and Singh, Amanpreet and Potts, Christopher and Kiela, Douwe and Mehri, Shikib},
  title = {Anchored Preference Optimization and Contrastive Revisions: Addressing Underspecification in Alignment},
  journal = {Transactions of the Association for Computational Linguistics},
  volume = {13},
  pages = {442-460},
  year = {2025},
  month = {04},
  issn = {2307-387X},
  doi = {10.1162/tacl_a_00748},
  url = {https://doi.org/10.1162/tacl\_a\_00748},
  eprint = {https://direct.mit.edu/tacl/article-pdf/doi/10.1162/tacl\_a\_00748/2522342/tacl\_a\_00748.pdf}
}

@article{pal2024smaug,
  title = {Smaug: Fixing Failure Modes of Preference Optimisation with {DPO}-Positive},
  author = {Pal, Arka and Karkhanis, Deep and Dooley, Samuel and Roberts, Manley and Naidu, Siddartha and White, Colin},
  journal = {Computing Research Repository (CoRR)},
  volume = {arXiv:2402.13228},
  year = {2024},
  url = {https://arxiv.org/abs/2402.13228}
}

@inproceedings{razin2025unintentional,
  title = {Unintentional Unalignment: Likelihood Displacement in Direct Preference Optimization},
  author = {Noam Razin and Sadhika Malladi and Adithya Bhaskar and Danqi Chen and Sanjeev Arora and Boris Hanin},
  booktitle = {The Thirteenth International Conference on Learning Representations},
  year = {2025},
  url = {https://openreview.net/forum?id=uaMSBJDnRv}
}

@article{chakrabarty2025ai,
  title = {{AI}-Slop to {AI}-Polish? Aligning Language Models through Edit-Based Writing Rewards and Test-Time Computation},
  author = {Chakrabarty, Tuhin and Laban, Philippe and Wu, Chien-Sheng},
  journal = {Computing Research Repository (CoRR)},
  volume = {arXiv:2504.07532},
  year = {2025},
  url = {https://arxiv.org/abs/2504.07532}
}

@inproceedings{ye-etal-2025-improving,
  title = {Improving Reward Models with Synthetic Critiques},
  author = {Ye, Zihuiwen  and
      Greenlee, Fraser David  and
      Bartolo, Max  and
      Blunsom, Phil  and
      Campos, Jon Ander  and
      Gall{\'e}, Matthias},
  editor = {Chiruzzo, Luis  and
      Ritter, Alan  and
      Wang, Lu},
  booktitle = {Findings of the Association for Computational Linguistics: NAACL 2025},
  month = apr,
  year = {2025},
  address = {Albuquerque, New Mexico},
  publisher = {Association for Computational Linguistics},
  url = {https://aclanthology.org/2025.findings-naacl.254/},
  pages = {4506--4520},
  isbn = {979-8-89176-195-7}
}

@inproceedings{pattnaik-etal-2024-enhancing,
  title = {Enhancing Alignment using Curriculum Learning {\&} Ranked Preferences},
  author = {Pattnaik, Pulkit  and
      Maheshwary, Rishabh  and
      Ogueji, Kelechi  and
      Yadav, Vikas  and
      Madhusudhan, Sathwik Tejaswi},
  editor = {Al-Onaizan, Yaser  and
      Bansal, Mohit  and
      Chen, Yun-Nung},
  booktitle = {Findings of the Association for Computational Linguistics: EMNLP 2024},
  month = nov,
  year = {2024},
  address = {Miami, Florida, USA},
  publisher = {Association for Computational Linguistics},
  url = {https://aclanthology.org/2024.findings-emnlp.754/},
  doi = {10.18653/v1/2024.findings-emnlp.754},
  pages = {12891--12907}
}

@article{xiao2025finding,
  title = {Finding the Sweet Spot: Preference Data Construction for Scaling Preference Optimization},
  author = {Xiao, Yao and Ye, Hai and Chen, Linyao and Ng, Hwee Tou and Bing, Lidong and Li, Xiaoli and Lee, Roy Ka-wei},
  journal = {arXiv preprint arXiv:2502.16825},
  year = {2025},
  url = {https://arxiv.org/abs/2502.16825}
}

@article{zhang2015character,
  title = {Character-Level Convolutional Networks for Text Classification},
  author = {Zhang, Xiang and Zhao, Junbo and LeCun, Yann},
  journal = {Advances in Neural Information Processing Systems},
  volume = {28},
  year = {2015},
  url = {https://arxiv.org/abs/1509.01626}
}

@article{just2025datacentric,
  title = {Data-Centric Human Preference with Rationales for Direct Preference Alignment},
  author = {Hoang Anh Just and Ming Jin and Anit Sahu and Huy Phan and Ruoxi Jia},
  year = {2025},
  archiveprefix = {arXiv},
  primaryclass = {cs.LG},
  journal = {arXiv preprint arXiv:2407.14477},
  url = {https://arxiv.org/abs/2407.14477}
}

@misc{wikidump2024,
  author = {{Wikimedia Foundation}},
  title = {Wikimedia Downloads},
  url = {https://dumps.wikimedia.org},
  year = 2024
}

@misc{alpaca_eval,
  author = {Xuechen Li and Tianyi Zhang and Yann Dubois and Rohan Taori and Ishaan Gulrajani and Carlos Guestrin and Percy Liang and Tatsunori B. Hashimoto },
  title = {{AlpacaEval}: An Automatic Evaluator of Instruction-Following Models},
  year = {2023},
  month = {5},
  publisher = {GitHub},
  journal = {GitHub repository},
  howpublished = {\url{https://github.com/tatsu-lab/alpaca_eval}}
}

@article{lin2024interpretable,
  title = {Interpretable User Satisfaction Estimation for Conversational Systems With Large Language Models},
  author = {Lin, Ying-Chun and Neville, Jennifer and Stokes, Jack W and Yang, Longqi and Safavi, Tara and Wan, Mengting and Counts, Scott and Suri, Siddharth and Andersen, Reid and Xu, Xiaofeng and others},
  journal = {arXiv preprint arXiv:2403.12388},
  year = {2024},
  url = {https://arxiv.org/abs/2403.12388}
}

@inproceedings{shi2024wildfeedback,
  title = {WildFeedback: Aligning {LLM}s With In-situ User Interactions And Feedback},
  author = {Taiwei Shi and Zhuoer Wang and Longqi Yang and Ying-Chun Lin and Zexue He and Mengting Wan and Pei Zhou and Sujay Kumar Jauhar and Xiaofeng Xu and Xia Song and Jennifer Neville},
  booktitle = {NeurIPS 2024 Workshop on Behavioral Machine Learning},
  year = {2024},
  url = {https://openreview.net/forum?id=07QCozT1pi}
}

@inproceedings{chowdhury2024provably,
  title = {Provably Robust {DPO}: Aligning Language Models With Noisy Feedback},
  author = {Chowdhury, Sayak Ray and Kini, Anush and Natarajan, Nagarajan},
  booktitle = {Proceedings of the 41st International Conference on Machine Learning},
  pages = {42258--42274},
  year = {2024},
  url = {https://arxiv.org/abs/2403.00409}
}

@article{huang2022selfimprove,
  title = {Large Language Models Can Self-Improve},
  author = {Jiaxin Huang and Shixiang Shane Gu and Le Hou and Yuexin Wu and Xuezhi Wang and Hongkun Yu and Jiawei Han},
  journal = {arXiv preprint arXiv:2210.11610},
  year = {2022},
  url = {https://arxiv.org/abs/2210.11610}
}

@article{guo2023leveraging,
  title = {Beyond Imitation: Leveraging Fine-grained Quality Signals for Alignment},
  author = {Geyang Guo and Ranchi Zhao and Tianyi Tang and Wayne Xin Zhao and Ji-rong Wen},
  journal = {arXiv preprint arXiv:2311.04072},
  year = {2023},
  url = {https://arxiv.org/abs/2311.04072}
}

@article{JMLR:v13:chiang12a,
  author = {David Chiang},
  title = {Hope and Fear for Discriminative Training of Statistical Translation Models},
  journal = {Journal of Machine Learning Research},
  year = {2012},
  volume = {13},
  number = {40},
  pages = {1159-1187},
  url = {http://jmlr.org/papers/v13/chiang12a.html}
}

@article{dong2024selfboosting,
  title = {Self-Boosting Large Language Models with Synthetic Preference Data},
  author = {Qingxiu Dong and Li Dong and Xingxing Zhang and Zhifang Sui and Furu Wei},
  journal = {arXiv preprint arXiv:2410.06961},
  year = {2024},
  url = {https://arxiv.org/abs/2410.06961}
}
\bibliographystyle{acl_natbib}

\appendix
\clearpage\newpage

\section{Prompts and Example Responses}
\label{appendix:prompts}

\paragraph{Query Generation Prompt.} The following prompt was provided to \texttt{gpt-4.1-2025-04-14}, with ``\{\{DOCUMENT\}\}'' replaced by a document from the target domain (e.g., Wikipedia, Yelp reviews, news articles, or arXiv papers), in order to generate queries.

\begin{displayquote}
    You're collaborating with a retrieval-augmented generation (RAG) system. The document below was retrieved in response to a user query.

Your task: Invent five \textbf{highly creative and cognitively demanding} queries that could have caused this document to be retrieved.

These queries should not just \textbf{use} the document---they should \textbf{stretch} it. Think of tasks that require deep reasoning, synthesis of ideas, subjective judgment, or unconventional thinking. Your goal is to make the model work hard.

Some examples:
\begin{itemize}
    \item ``Propose a novel startup idea grounded in the methodologies of this academic paper, and outline a go-to-market strategy.''
    \item ``Evaluate the emotional tone across customer reviews for this hotel and design a retraining program for staff based on your findings.''
\end{itemize}

Guidelines:
\begin{itemize}
    \item The user did \textbf{not} know the contents of this document before issuing the query.
    \item Each query must \textbf{necessitate} retrieving and deeply using the content of this document.
    \item \textbf{Be original}: No two queries should sound alike. Avoid formulaic patterns or generic phrasing.
    \item Simple questions, trivia-style queries, or those answerable in a few words are \textbf{not acceptable}.
    \item \textbf{Creativity is rewarded. Redundancy is penalized.}
\end{itemize}

Return your output as valid JSON in the following format:
{"queries": ["...", "...", "...", "...", "..."]}

\#\# Document

\{\{DOCUMENT\}\}
\end{displayquote}

\begin{figure*}[ht]
    \centering
    \includegraphics[width=\textwidth]{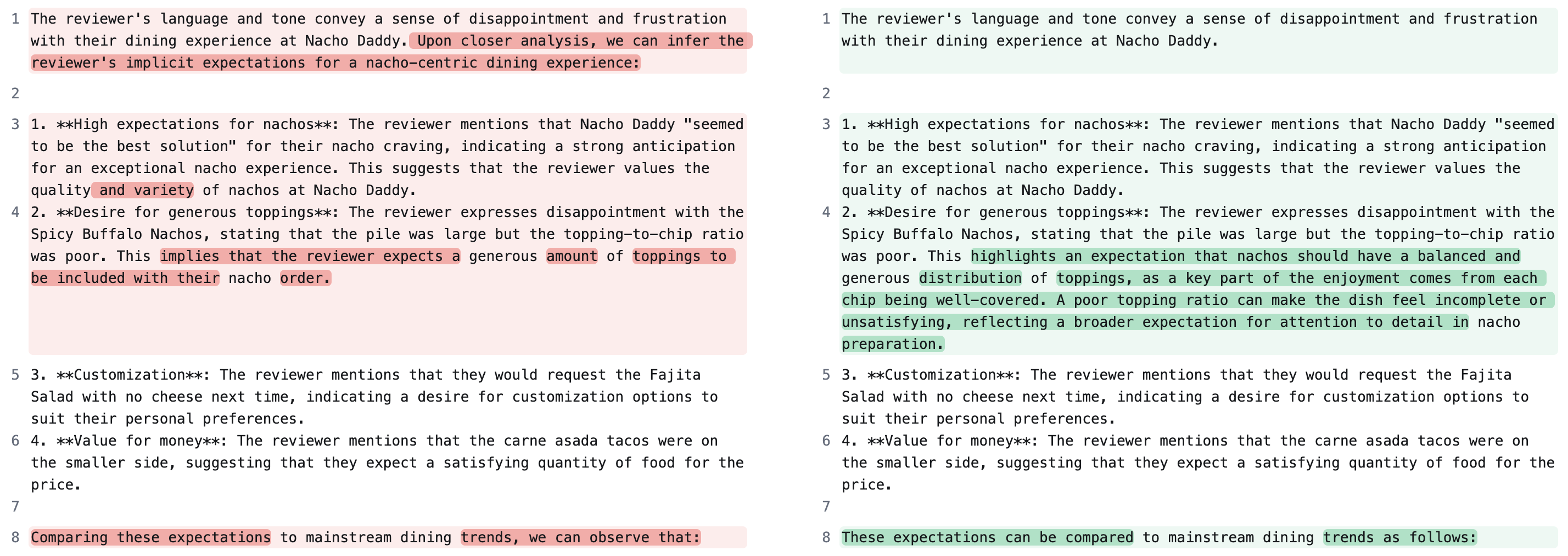}
    \caption{\textbf{Left}: the original response.
    \textbf{Right}: the fully edited response $a^*$ after incorporating all user-highlighted feedback. Differences are highlighted.}
    \label{fig:response_examples}
\end{figure*}

\paragraph{Query Examples.} The queries below are a random sample illustrating the types of queries generated for each domain.

\begin{displayquote}
    \textbf{Example 1} (arXiv): Analyze how the design choices in MERLOT Reserve regarding time alignment and masking strategies for audio and text affect the model's ability to learn temporal commonsense knowledge, and propose an ablation study to isolate these effects.

    \textbf{Example 2} (Wikipedia): Analyze how the spontaneous social interactions on beep lines prefigured the emergence of online social networks, identifying unique features and limitations. Compose a comparative critique highlighting lessons for the design of future virtual communities.

    \textbf{Example 3} (Yelp): Drawing inspiration from the review's depiction of customer service and venue ambiance, design a staff training program that balances `sass' with professionalism for quick-service restaurants.

    \textbf{Example 4} (News): Using the methodologies and results described in the study, propose a hypothetical follow-up experiment that would distinguish between the effects of pet exposure and other confounding factors, such as rural versus suburban environments or breastfeeding rates, on infant health outcomes.
\end{displayquote}

\paragraph{Response Generation Prompt.} The following prompt scaffold was provided to \texttt{meta-llama/\allowbreak Llama-3.1-8B-Instruct} to generate responses to the queries.

\begin{displayquote}
You are a helpful and intelligent assistant. Directly answer the user query while using information in the document provided.

\#\# Document

\{\{DOCUMENT\}\}

\#\# Query

\{\{QUERY\}\}
\end{displayquote}

\paragraph{Response Examples.} Figure~\ref{fig:response_examples} shows a partial example of an original response generated by \texttt{meta-llama/\allowbreak Llama-3.1-8B-Instruct} and a fully rewritten version that reflects the cumulative result of all step-wise edits addressing user feedback.

\paragraph{Response Rewrite Prompt.} The following system prompt was provided to \texttt{meta-llama/\allowbreak Llama-3.1-8B-Instruct} to generate step-wise cumulative improvement sequences. Dislike IDs provided are randomized.

\begin{displayquote}
    You are tasked with improving the following generated response based on user feedback, which includes highlighted likes and dislikes. Your goal is to produce a series of incremental and cumulative edits that improve the response by directly addressing the user’s dislikes.

You will be provided with the following:

    User Query: The original request from the user.
    Knowledge Source: A collection of relevant information retrieved by the system to support the response.
    Generated Response with User Feedback Highlights: The final output produced by a large language model based on the query and retrieved knowledge, with highlighted sections based on user feedback.
    User Feedback Highlight Reasons: The specific reasons provided by the user explaining their likes and dislikes for each highlighted portion of text. These reasons apply only to the highlighted segments, not to the response as a whole.

Your instructions:
    Improve the response in a sequence of steps.
    In each step, address \textbf{only one} <dislike> span based on its corresponding feedback.\jason{instructions don't say much to do about like spans -- shouldn't we say "don't edit away what made the response good"? -- and doesn't mention that like spans may overlap with dislike spans}
    You \textbf{must} output the entire edited response at each step—not just the edited portion.
    In each step, \textbf{remove the <dislike> highlight tag you are addressing from the text.}
    Highlight tags for \textbf{unaddressed dislikes must be preserved} in the edited text.
    You may revise surrounding text if needed to maintain coherence, flow, or tone.
    Carefully \textbf{check and fix any formatting errors} at each step (e.g., incorrect numbering, bullet points used incorrectly, inconsistent list formatting).
    \textbf{Avoid expanding or shrinking} the overall length of the response more than necessary. Try to keep each edit roughly the same length as the original text unless the feedback clearly calls for a change.
    \textbf{Do not over-optimize for conciseness} if it comes at the expense of fully addressing the feedback.
    Prioritize directly addressing the user's feedback meaningfully over unnecessary rewording.
    Important: Make sure to do the edits in the \textbf{ORDER OF DISLIKE IDS}. That is, start with ID = 1, then proceed to ID = 2, and so on.
\end{displayquote}

The user prompt given with this system prompt is structured as follows:

\begin{displayquote}
    \#\# User Query 

    \{\{User Query\}\}

    \#\# Knowledge Source

    \{\{Knowledge Source\}\}

    \#\# Generated Response with User Feedback Highlights

    \{\{Response with Feedback Highlights\}\}

    \#\# User Feedback Highlight Reasons

    \{\{Highlight Reasons\}\}
\end{displayquote}

An example displaying the formatting of the response with highlights together with the feedback reasons that are fed into the prompt above are as follows:\jason{can you use 2 spans instead of one, so that it's clear that all of the explanations come after the entire highlighted document?}

\begin{displayquote}
    ...Firstly, the document highlights the half-price offer after 11 PM, which is particularly appealing to college students. This pricing strategy allows students to affordably enjoy a meal, making it a staple for freshmen and beyond. <like id='1'>\textbf{The document states, "I started coming here, as all Pitt students do, freshman year, for the luxury of half-price food after 11." This suggests that the economic benefit of the half-price offer has fostered loyalty and encouraged repeat visits among students.}</like id='1'>...

    ...\{`id': 1, `explanation': ``I like this because it states a useful fact. I like this because it quotes/cites/paraphrases from the retrieved document.''\}
\end{displayquote}

\section{Feedback Taxonomy}
\label{appendix:feedback}

\subsection{Feedback on \textbf{like} spans}

\begin{itemize}
    \item Utility: I like this because...
        \begin{itemize}
            \item It directly answers my question.
            \item It smoothly leads up to answering my question.
            \item It helps my general understanding of the topic.
            \item It gives a quick recap/summary.
        \end{itemize}
    \item Where the information comes from: I like this because...
    \begin{itemize}
        \item It quotes/cites/paraphrases from the retrieved document.
        \item It echos/repeats/reiterates my query.
        \item The assistant generated it on its own.
    \end{itemize}
    \item What the span contributes: I like this because...
    \begin{itemize}
        \item It states a useful fact.
        \item It draws a sensible conclusion.
        \item It assesses how reliable the info is.
        \item It suggests one or more possibilities (e.g., examples, options, explanations, considerations).
        \item It defines a term or explains a concept.
        \item It offers an opinion.
        \item It corrects or clarifies my question/instructions.
        \item It flags an important caveat or potential pitfall.
        \item It acknowledges a limitation or the uncertainty involved.
    \end{itemize}
    \item Craft \& style: I like this because...
    \begin{itemize}
        \item It shows careful attention to the details of my query.
        \item It's well-written.
        \item It's well-organized.
        \item It’s engaging—maybe even a little funny.
    \end{itemize}
    \item Other: Describe why you liked this span.
\end{itemize}

\subsection{Feedback on \textbf{dislike} spans}

\begin{itemize}
    \item Poor content: I dislike this because...
    \begin{itemize}
        \item It states something that’s factually wrong.
        \item The opinion or advice it gives is weak or low-quality.
        \item It doesn’t add anything useful—totally unnecessary.
        \item It’s off-topic or irrelevant to my question.
        \item The wording is toxic or offensive.
    \end{itemize}

    \item Misleading sourcing: I dislike this because...
    \begin{itemize}
        \item It credits the wrong source for a claim.
        \item It misrepresents what the source actually says.
    \end{itemize}

    \item Craft \& style problems: I dislike this because...
    \begin{itemize}
        \item It dumps too many options on me at once.
        \item The writing is confusing or hard to follow.
        \item It repeats itself unnecessarily.
        \item It’s too wordy.
        \item The tone or style doesn’t fit what I expect.
    \end{itemize}

    \item Other issues: I dislike this because...
    \begin{itemize}
        \item It feels generic or incomplete.
        \item It shows the assistant misunderstood my question or instructions.
        \item It ignores the instructions I gave.
        \item It’s too one-sided and misses other perspectives.
        \item It lacks depth or useful detail.
        \item It just copies the source without adding insight.
        \item I just disagree with it.
    \end{itemize}
    \item Other: Describe why you disliked this span.
\end{itemize}

\begin{figure*}[ht]
    \centering
    \includegraphics[width=0.8\textwidth]{images/annotation_interface.png}
    \caption{Span highlighting and rationale providing annotation interface, showing the spans highlighted by a user and the provided reasons for those highlights.}
    \label{fig:annotation_interface}
\end{figure*}

\section{Annotation Guidelines and Interface}
\label{appendix:annotation}

\paragraph{Preamble.} For each annotation item, first take some time to read the Query. Imagine that you’ve posed the following query to an AI assistant that will search the internet for related documents and webpages to answer your query. Think about what you would like to see in a given response to this query. Next, briefly skim the Knowledge Source. Suppose that this is the document the assistant found on the internet to base its response to your query off of. The information present in this document and the knowledge present in the language model behind the assistant will be the only information the assistant can use to answer your query. Given this, think about what you would like to see in a given response to the query, that is based on the knowledge present in this knowledge source document.

\paragraph{Span Highlighting.} You are presented with two responses to the query above, both of which are based on the information present in the knowledge source and the knowledge present in the large language model underlying an AI assistant. Read each response carefully. Your task is to identify and highlight any sections of text that you liked or disliked in each response. To highlight a span, simply drag your cursor over the specific text itself; you can change your current highlight mode (like or dislike) by clicking on the Like Highlight Mode / Dislike Highlight Mode button switcher at the top. Spans you like will be highlighted in green; red for dislike.

When you finish highlighting a single span, a popup window will appear asking you why you liked or disliked the span you highlighted. Read through the options, ticking the corresponding checkbox(es) if the reason(s) correspond to why you liked or disliked the span you just highlighted. You can tick as many checkboxes as you want here; if there are any other reasons in addition to those in the checkbox that made you like the given span, please tick the final Other checkbox and briefly describe your reason in the corresponding text box. Press Save when you’re finished highlighting the reasons behind your highlight. 

You can revisit and edit your answers to each checkbox easily by simply clicking on the span you highlighted, which will bring up your answers you chose for the corresponding span. The span highlight will darken slightly in color, to indicate that the window popup currently corresponds to the indicated span.

Please make these highlights for both Response A and Response B. When deciding how large or how small a span to highlight, imagine that you were directing someone's attention to the fact that you liked or disliked a given span. \textit{Which parts, if any, for each would you highlight, for another to check this as quickly as possible?}

\paragraph{Overlapping Spans.} You may like and dislike a part of or a given span for different reasons—for example, you may like the content of a span, but dislike its style or how the information is presented. In such cases, it’s encouraged to note this down by having overlapping spans either partially or completely overlapping spans of text; simply select the opposite highlight mode and highlight over an already highlighted span of text.

\paragraph{Response-Level Questions.} Once you’ve finished highlighting spans of text that you liked and disliked in both responses, you are now to highlight why you may have liked or disliked each given response as a whole. Tick off the corresponding checkboxes if you feel that they apply to each given response, and write out any reasons why you may have liked or disliked a given response as a whole for reasons not present in the checkboxes given.

\paragraph{AB Preference Ranking.} Finally, think about both responses, and the annotations you made. Make a decision—which response did you prefer more? Indicate your choice, and write down why you did so, in the provided space below. \textit{Important: The tie option should be used very, very rarely—reserve it for the exceptional cases where the two responses are genuinely indistinguishable (or equally poor). Try to make a decisive and justified decision.}

\paragraph{Annotation Interface.} Figure~\ref{fig:annotation_interface} illustrates the annotation interface, displaying two responses side by side. In the example, a user has selected a span in Response B. The example reveals the specific reasons the user provided for disliking the span.

\section{Details}
\label{appendix:hyperparams}

\paragraph{Hyperparameters.} Direct alignment models were post-trained with a maximum sequence length of 8192 tokens and a global batch size of 4. Optimization was performed using AdamW with a learning rate of 5e-7, scheduled via cosine decay and no warm-up. Mixed-precision (fp16) training and gradient checkpointing were enabled, and a DPO $\beta$ value of 0.1 was used. SFT models were fine-tuned with a learning rate of 5e-6. Grid search on the learning rate was performed for both SFT and direct alignment models over the values of 5e-6, 2e-6, 1e-6, 5e-7, and 2e-7.

\paragraph{Elo Score Calculation.} Pairwise comparison counts (\textit{wins} = 1, \textit{draws} = 0.5, \textit{losses} = 0) were expanded into individual game outcomes and fitted with a Bradley-Terry (logistic Elo) model, where the probability that model $i$ defeats model~$j$ is 
\[
P_{ij} \;=\; \frac{1}{1 + 10^{(R_j - R_i)/400}} .
\]
Ratings $R_i$ were obtained by maximizing the joint log-likelihood.
As individual A/B judgments are not IID, we repeatedly drew 1,000 bootstrap samples by resampling the set of prompts---keeping all A/B judgment-level outcomes among its responses---with replacement, refit the Elo model on each sample, and took the 2.5th/97.5th percentiles of the resulting \textit{pair-wise rating differences} as the 95\% confidence limits.

\end{document}